\pdfoutput=1

\documentclass[11pt]{article}

\usepackage[]{emnlp2021}

\usepackage{times}
\usepackage{latexsym}

\usepackage[T1]{fontenc}

\usepackage[utf8]{inputenc}
\usepackage{booktabs}
\usepackage{multirow}
\usepackage{graphicx}

\usepackage{microtype}

\title{Investigating Numeracy Learning Ability of a Text-to-Text Transfer Model}

\author{
Kuntal Kumar Pal,
Chitta Baral\\
{Department of Computer Science}\\
{Arizona State University, Tempe, Arizona, USA},\\
kkpal@asu.edu, chitta@asu.edu}

\begin{document}
\maketitle
\begin{abstract}

The transformer-based pre-trained language models have been tremendously successful in most of the conventional NLP tasks. But they often struggle in those tasks where numerical understanding is required. Some possible reasons can be the tokenizers and pre-training objectives which are not specifically designed to learn and preserve numeracy. Here we investigate the ability of  text-to-text transfer learning model (T5), which has outperformed its predecessors in the conventional NLP tasks, to learn numeracy. We consider four numeracy tasks : numeration, magnitude order prediction, finding minimum and maximum in a series, and sorting. We find that, although T5 models perform reasonably well in the interpolation setting, they struggle considerably in the extrapolation setting across all four tasks.
\end{abstract}

\section{Introduction}






Recent advances in transfer learning in NLP have led to the emergence of pre-trained models which show a much stronger contextual representation of words than earlier static word embeddings. They have all performed extremely well in conventional NLP tasks. Yet, they fail to capture a better understanding of numbers. 
Numbers are integral part of natural language texts which can change the meaning of a sentence. 
So there is a need for NLP models which can identify numbers represented in any surface forms like words, floats or strings (Numeration), understand its values in various context (Magnitude Order Prediction), compare their values with others (List-MinMax) or able to rearrange a series of numbers based on its values (Sorting).

The transfer-learned models are pre-trained on huge amount of natural language texts with specially designed tasks and tokenizers to create stronger word-embeddings. This causes the numbers embedded in the texts to lose their meaning and inherent rules of numeracy guiding them \cite{thawani2021representing,nogueira2021investigating}. This is possibly the reason they perform worse in numerical reasoning tasks on numbers absent in training data \cite{nogueira2021investigating,wallace-etal-2019-nlp}.

\begin{figure}
  \includegraphics[width=\columnwidth]{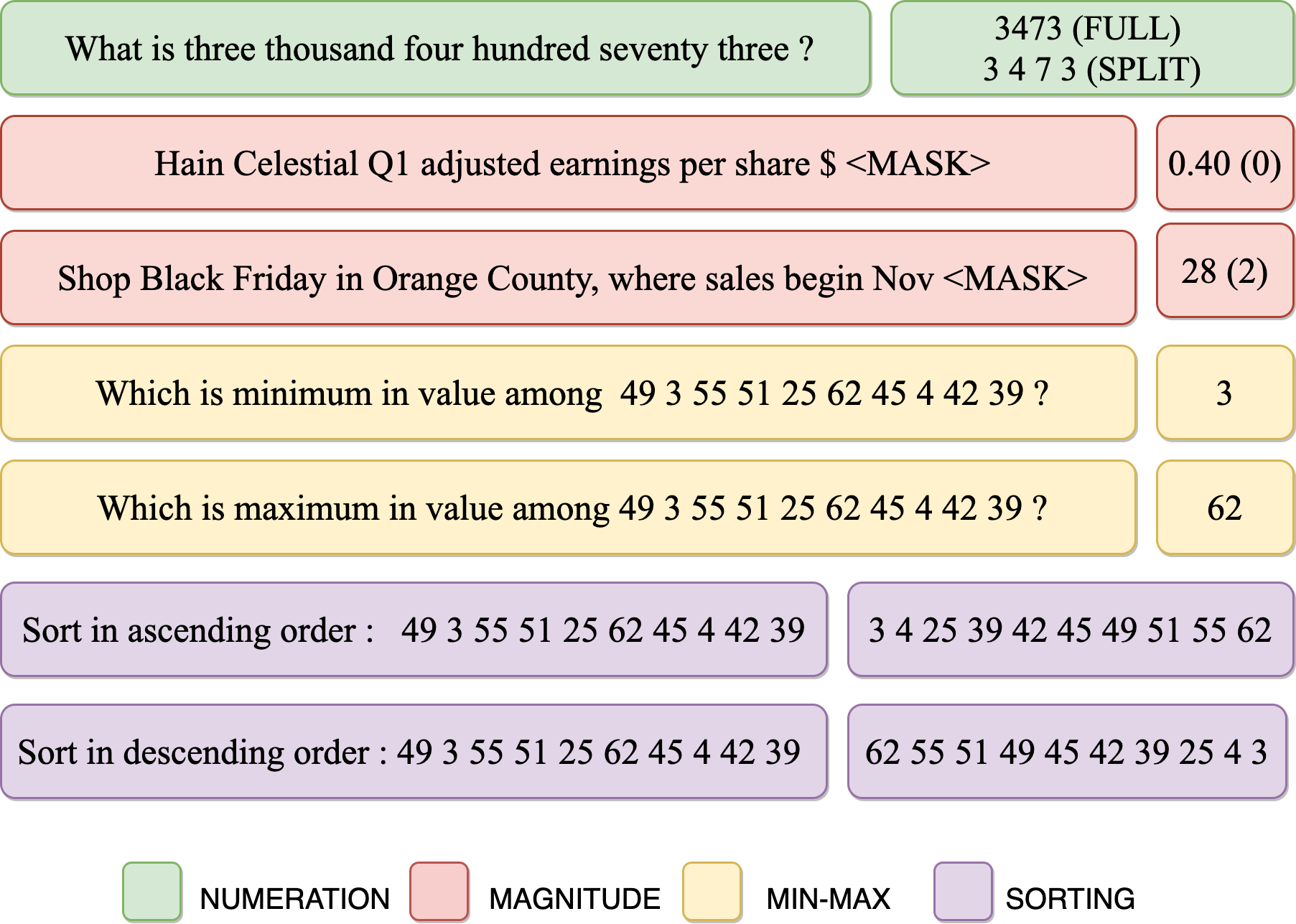}
  \caption{Examples of Numeracy Tests}
  \label{fig:tasks}
\end{figure}

In this paper, we test this numeracy learning ability of  a text-to-text transfer learning generative model, T5 \cite{T5} which has outperformed its predecessors in conventional NLP tasks. 
The text-to-text format of input and output helps the model to generalize all the NLP tasks as a unified model. We use  four numeracy tests both in interpolation (training and testing on same range of data) and extrapolation settings (training on lower and testing on higher range of data) and study how much numeracy skill it can acquire. Figure \ref{fig:tasks} shows some examples of each of the numeracy tests.

Our contributions in this paper are: (1) Extensive study on three versions of T5 models (small, base, large) on four numeracy tests in interpolation and extrapolation settings. 
(2) Reporting interesting observations in the behavior of each model version across multiple experimental settings  through  detailed manual error analysis.
The synthetically generated data and codes are publicly available\footnote{https://github.com/kuntalkumarpal/T5Numeracy} for future numeracy analysis in similar settings.

\section{Numeracy Tests}

We perform four essential numeracy tests to explore model's  ability to understand  numerical values. 

\noindent
\textbf{Motivation:} These four elementary  tasks are simple and easy for the models, since they do not need to generate a completely new number in a different numerical range (like in mathematical tests : multiplication, division, exponentiation). Here we evaluate whether the models learn the numeracy tasks or they simply learn bias from the number range seen in training data.

\subsection{Numeration}
The probability of a number represented in multiple surface forms (word, scientific, float, integer)  increases with the increase in the volume of  pre-training corpus of the language models. It is impractical for an end-to-end NLP model to semantically parse these numbers accurately and convert into a single representation to retain its value or reason with. This task tests the model's ability to understand word representation of a number and to decode into integer form.


\subsection{Magnitude Order Prediction}
The task is to identify the order of magnitude of a missing (masked) number which fits the context of a natural language text. This task is important in numerical commonsense reasoning \cite{lin2020numersense} and prompt-based methods \cite{liu2021pretrain}. Here, we do not expect the model to predict the  exact number that fits the context as this may vary in different domains. 
Instead, this task  tests the model's ability to understand a missing number's context and predict its appropriate range.

\subsection{List-MinMax}
We test the model's ability to understand numerical values and compare among them. Given a series of $n$ positive numbers, the task is to find the minimum and the maximum number. This is the basis of many question answering and commonsense numerical reasoning dataset like SQuAD \cite{rajpurkar-etal-2016-squad}, DROP \cite{dua-etal-2019-drop} and NUMBERGAME \cite{mishra2020towards}.
We simplify the task by generating templates  so that the models can concentrate on understanding the task rather than getting confused by the language complexities.

\subsection{Sorting}
In addition to understanding the values of each number in a series, the model will have to rearrange them in the correct order through this task, making it even harder than  List-MinMax. Even if a model is successful in the previous test, it is necessary to identify whether it has actually compared among all the numbers in the series. Hence, sorting a list of $n$ numbers in ascending and descending orders ensures that the model compares all the numbers and rearrange them into two different sequences.

\section{Experiments}

\noindent
\subsection{Experimental Setup:}
We use T5-SM (small, 60M parameters), T5-BS (base, 220M), T5-LG (large, 770M)  and  positive integers for the experiments. The results are average of three random seeds.
We perform experiments in two settings: \textit{interpolation}  (training  and  testing  on  same  numerical range) and \textit{extrapolation} (training on lower and testing on higher numerical range). The latter helps us to analyze whether a model has learnt the task, or it has exploited bias in the numerical range of the training data.

\subsection{Data Preparation:}

\begin{table}[]
\centering
\resizebox{\columnwidth}{!}{%
\begin{tabular}{@{}rrrrrrrr@{}}
\toprule
\multicolumn{2}{l}{\textbf{\# TRAIN  $\rightarrow{}$}} &
  \multicolumn{2}{c}{\textbf{4.9K}} &
  \multicolumn{2}{c}{\textbf{1.3K}} &
  \multicolumn{2}{c}{\textbf{0.9K}} \\ \cmidrule(l){3-8}
\multicolumn{1}{l}{\textbf{TP} } &
  \textbf{Model}  &
  \multicolumn{1}{c}{\textbf{IN}} &
  \multicolumn{1}{c}{\textbf{EX}} &
  \multicolumn{1}{c}{\textbf{IN}} &
  \multicolumn{1}{c}{\textbf{EX}} &
  \multicolumn{1}{c}{\textbf{IN}} &
  \multicolumn{1}{c}{\textbf{EX}} \\
  \midrule
\multirow{3}{*}{FL}  & T5-SM &  45.31 & 0.08 & 1.90 & 0.01 & 0.33 & 0.00 \\
                        & T5-BS &  92.16 & 1.03 & 66.47 & 0.45 & 37.20 & 0.42 \\
                        & T5-LG &  98.06 & 1.91 & 89.49 & 1.96 & 79.48 & 1.58 \\
                        \midrule
                        
\multirow{3}{*}{SP}   & T5-SM & 69.67 & 39.35 & 26.89 & 1.10 & 0.23 & 0.01  \\
                        & T5-BS &  99.50 & 11.31 & 81.21 & 22.44 & 73.61 & 31.06  \\
                        & T5-LG &  100.00 & 10.05 & 99.97 & 7.35 & 91.59 & 12.92  \\
\bottomrule
\end{tabular}%
}
\caption{\textbf{Numeration} EM scores w/ split (SP) and w/o split (FL) representation on  4.9K, 1.3K, 0.9K train-data in Interpolation (IN) and Extrapolation (EX) settings.}
\label{tab:decoding}
\end{table}

\begin{table*}[]
\centering
\resizebox{12cm}{!}{%
\begin{tabular}{@{}rrrrrrrr|rrrrrr@{}}
\toprule
& &
\multicolumn{6}{c}{\textbf{\uppercase{List Minimum} }} &\multicolumn{6}{c}{\textbf{\uppercase{List Maximum} }} \\
\midrule
\multicolumn{2}{l}{\textbf{\# ELEMENTS }} &
  \multicolumn{2}{c}{\textbf{3}} &
  \multicolumn{2}{c}{\textbf{5}} &
  \multicolumn{2}{c}{\textbf{10}}&
  \multicolumn{2}{c}{\textbf{3}} &
  \multicolumn{2}{c}{\textbf{5}} &
  \multicolumn{2}{c}{\textbf{10}} \\ \cmidrule(l){3-14}
\multicolumn{1}{l}{\textbf{Range} } &
  \textbf{Model}  &
  \multicolumn{1}{c}{\textbf{IN}} &
  \multicolumn{1}{c}{\textbf{EX}} &
  \multicolumn{1}{c}{\textbf{IN}} &
  \multicolumn{1}{c}{\textbf{EX}} &
  \multicolumn{1}{c}{\textbf{IN}} &
  \multicolumn{1}{c}{\textbf{EX}}  &
  \multicolumn{1}{c}{\textbf{IN}} &
  \multicolumn{1}{c}{\textbf{EX}} &
  \multicolumn{1}{c}{\textbf{IN}} &
  \multicolumn{1}{c}{\textbf{EX}} &
  \multicolumn{1}{c}{\textbf{IN}} &
  \multicolumn{1}{c}{\textbf{EX}}\\
  \midrule
\multirow{3}{*}{< 99}   & T5-SM & 90.5 & 0.6 & 86.5 & 0.1 & 65.9 & 0.0 &  80.4 & 0.5 & 71.6 & 0.3 & 74.7 & 0.1 \\
                        & T5-BS & 96.2 & 33.9 & 99.1 & 13.0 & 98.2 & 2.8 &  92.3 & 22.7 & 96.8 & 6.0 & 90.4 & 1.1\\
                        & T5-LG & 100.0 & 22.2 & 99.4 & 2.8 & 100.0 & 0.5 &  100.0 & 29.6 & 100.0 & 13.6 & 100.0 & 2.0 \\
                        \midrule
\multirow{3}{*}{< 999}  & T5-SM &  72.6 & 41.8 & 55.5 & 22.2 & 49.9 & 9.7 &  65.3 & 38.4 & 54.8 & 17.5 & 40.0 & 5.2\\
                        & T5-BS &  91.5 & 67.2 & 92.1 & 42.6 & 80.4 & 27.1 & 89.1 & 65.3 & 90.8 & 47.2 & 88.3 & 25.0\\
                        & T5-LG & 98.3 & 70.1 & 96.1 & 49.3 & 87.4 & 34.7 &  96.1 & 61.2 & 97.8 & 58.7 & 95.2 & 35.3\\
                        \midrule
\multirow{3}{*}{< 9999} & T5-SM & 59.1 & 44.7 & 43.5 & 30.4 & 30.7 & 17.1 &  51.2 & 47.0 & 36.0 & 27.0 & 20.9 & 11.1\\
                        & T5-BS & 89.6 & 68.8 & 86.9 & 53.8 & 85.4 & 38.1 &  87.1 & 58.6 & 83.1 & 43.4 & 81.6 & 29.9\\
                        & T5-LG & 97.1 & 81.3 & 93.7 & 71.8 & 94.0 & 58.2 &  96.2 & 84.9 & 94.9 & 76.4 & 94.9 & 59.1 \\
\bottomrule
\end{tabular}
}
\caption{\textbf{List-MinMax} (series  length: 3, 5, 10) in three different number ranges evaluated as Interpolation (IN) and Extrapolation (EX) exact-match scores on 1K test data.}
\label{tab:minmax}
\end{table*}

\noindent
\textbf{Numeration:} We create a dataset keeping in mind that at least few examples of all unique words needed to represent each number, are present in the training data \cite{trasknalu}. In Table \ref{tab:decoding}, interpolation samples are from [0,10K) and 99K extrapolation samples are from [10K,1000K). We use \textit{num2words}\footnote{https://github.com/savoirfairelinux/num2words} for generating word-form of each integer. To simulate fewer shot setting, we carefully craft two smaller training sets taking only 20\% and 10\% data. We show two number representation schemes with split-digits (SP) and without split (FL)  hypothesizing that for a generative model it would be easier to correctly generate individual digits instead of full integer at once.

\begin{table}[]
\centering
\small
\begin{tabular}{@{}lcccc@{}}
\toprule
\textbf{Datasets   $\rightarrow$ }       & \multicolumn{2}{c}{\textbf{AT}} & \multicolumn{2}{c}{\textbf{MC}} \\ \cmidrule{2-5}
\textbf{Models $\downarrow$}          & \textbf{$\mu$F1}              & \textbf{$m$F1}              & \textbf{$\mu$F1}            & \textbf{$m$F1}            \\

\midrule
LR               &62.49 & 30.81 & 71.25 & 60.80 \\ 
CNN              &69.27 & 35.96 & 77.17 & 58.49 \\
GRU              &70.92 & 38.43 & 78.25 & 58.08 \\ 
BiGRU            &71.49 & 39.94 & \underline{80.16} & 62.74 \\
CRNN             &69.50 & 36.15 & 78.00 & \underline{64.62} \\ 
CNN-capsule      & 63.11 & 29.41 & 75.89 & 59.22 \\ 
GRU-capsule      &70.73 & 33.57 & 77.36 & \textbf{64.71} \\ 
BiGRU-capsule    &71.49 & 34.18 & 77.97 & 64.34 \\ 
BiLSTM-DICE      &75.56 & \textbf{46.80} &  - & - \\ 
\midrule
T5-SM            &69.87 & 31.36 & 66.11 & 34.68 \\ 
T5-BS            & \underline{78.06} & 40.04 & 72.22 & 47.44 \\
T5-LG            &  \textbf{81.40} & \underline{44.64} & \textbf{80.29} & 59.16 \\
\bottomrule
\end{tabular}

\caption{\textbf{Magnitude Order Prediction} for Market Comments (MC) and Article Titles (AT) datasets of numeracy600K in micro-F1 ($\mu$F1) and macro-F1 ($m$F1). Best score is in bold and second-best is underlined. } 
\label{tab:magnitude}

\end{table}
\noindent
\textbf{Magnitude Order Prediction:} For this task we work on Numeracy600K \cite{chen2019numeracy} dataset. We consider this as a mask prediction  task. We train  models to find the exact number that fits the mask. Then, we map the predicted numbers into its magnitude order, save the  model based on best magnitude order and calculate the evaluation metrics on test data. Since this is a generation task we reject those answers which are not valid floating point numbers. 
The baseline results in Table \ref{tab:magnitude} are  from \cite{chen2019numeracy,sundararaman2020methods}. We also consider extrapolation setting by showing the cross-domain performance (train on market comments and test on article title  and vice-versa) in Table \ref{tab:magnitude_cross}. 

\begin{table}[]
\centering
\small
\begin{tabular}{@{}lllll@{}}
\toprule\textbf{Train on $\rightarrow$} & \multicolumn{2}{c}{\textbf{AT}} & \multicolumn{2}{c}{\textbf{MC}} \\ \cmidrule{2-5}
\textbf{Models $\downarrow$}  & \textbf{$\mu$F1}            & \textbf{$m$F1}       & \textbf{$\mu$F1}            & \textbf{$m$F1}       \\
\midrule
BiGRU &            25.59 & 10.58 & 31.38 & 11.08 \\
\midrule
T5-SM       &          28.88 & 12.04 & 37.35 & 10.81 \\ 
T5-BS       &          35.53 & 14.48 & 31.51 & 12.25 \\ 
T5-LG       &          \textbf{50.18} & \textbf{21.24} & \textbf{38.43} & \textbf{12.32}  \\  \bottomrule
\end{tabular}%
\caption{\textbf{Cross Domain} (Extrapolation) Tests of  Order Prediction. Train on MC, test on AT and vice-versa. } 
\label{tab:magnitude_cross}
\end{table}

\noindent
\textbf{List Min-Max \& Sort:} We experiment on three different number ranges: [0,100), [0,1K), [0,10K). For interpolation tests, the numbers in the test data are from the same ranges. The extrapolation numbers are from the maximum of respective ranges to 100K. To prevent the model's bias on number lengths, we bring them closer following prior work \cite{wallace-etal-2019-nlp}. We extend the experiment on a series of 3, 5 and 10 numbers (for each range) to study how each of the models behave with increasing series length. We consider the same data for sorting experiments as well.
    The results are in Table \ref{tab:minmax} for List-MinMax and Table \ref{tab:sort} for List-Sort.

\begin{table*}[]
\centering
\resizebox{12cm}{!}{%
\begin{tabular}{@{}rrrrrrrr|rrrrrr@{}}
\toprule
& & 
\multicolumn{6}{c}{\textbf{\uppercase{List-Sort Ascending }}} & 
\multicolumn{6}{c}{\textbf{\uppercase{List-Sort Descending }}}\\
\midrule
\multicolumn{2}{l}{\textbf{\# ELEMENTS}} &
  \multicolumn{2}{c}{\textbf{3}} &
  \multicolumn{2}{c}{\textbf{5}} &
  \multicolumn{2}{c}{\textbf{10}} &
  \multicolumn{2}{c}{\textbf{3}} &
  \multicolumn{2}{c}{\textbf{5}} &
  \multicolumn{2}{c}{\textbf{10}}\\ \cmidrule(l){3-14}
\multicolumn{1}{l}{\textbf{Range} } &
  \textbf{Model}  &
  \multicolumn{1}{c}{\textbf{IN}} &
  \multicolumn{1}{c}{\textbf{EX}} &
  \multicolumn{1}{c}{\textbf{IN}} &
  \multicolumn{1}{c}{\textbf{EX}} &
  \multicolumn{1}{c}{\textbf{IN}} &
  \multicolumn{1}{c}{\textbf{EX}} &
  \multicolumn{1}{c}{\textbf{IN}} &
  \multicolumn{1}{c}{\textbf{EX}} &
  \multicolumn{1}{c}{\textbf{IN}} &
  \multicolumn{1}{c}{\textbf{EX}} &
  \multicolumn{1}{c}{\textbf{IN}} &
  \multicolumn{1}{c}{\textbf{EX}} \\
  \midrule
\multirow{3}{*}{< 99}   & T5-SM & 54.0 & 12.4 & 7.6 & 0.0 & 0.0 & 0.0 & 56.0 & 12.6 & 5.9 & 0.4 & 0.0 & 0.0 \\
                        & T5-BS &  80.6 & 12.2 & 87.2 & 0.0 & 0.4 & 0.0 & 84.3 & 12.9 & 75.5 & 0.0 & 6.2 & 0.0 \\ 
                        & T5-LG &  100.0 & 5.8 & 99.9 & 0.0 & 69.7 & 0.1 & 100.0 & 13.1 & 96.6 & 0.1 & 57.6 & 0.1\\ 
                        \midrule
\multirow{3}{*}{< 999}  & T5-SM & 32.6 & 15.1 & 1.4 & 0.6 & 0.0 & 0.0 & 38.0 & 22.3 & 3.4 & 1.3 & 0.0 & 0.0\\ 
                        & T5-BS & 74.7 & 45.7 & 64.0 & 8.0 & 12.5 & 0.0 &  73.1 & 42.0 & 62.6 & 9.6 & 16.8 & 0.1\\ 
                        & T5-LG &  95.1 & 64.2 & 91.8 & 16.8 & 61.9 & 1.7 & 94.7 & 63.5 & 92.5 & 25.7 & 61.2 & 1.6\\
                        \midrule 
\multirow{3}{*}{< 9999} & T5-SM &  23.4 & 17.1 & 1.0 & 0.1 & 0.0 & 0.0 &  30.4 & 21.2 & 0.7 & 0.4 & 0.0 & 0.0\\ 
                        & T5-BS &  63.1 & 45.5 & 51.1 & 12.7 & 15.0 & 0.2 & 59.8 & 43.9 & 51.4 & 12.4 & 14.3 & 0.3\\ 
                        & T5-LG & 94.5 & 76.0 & 87.4 & 43.2 & 74.6 & 12.6 & 94.2 & 76.1 & 86.1 & 44.4 & 75.6 & 11.9\\ 
\bottomrule
\end{tabular}
}


\caption{ \textbf{List-Sort (Ascending \& Descending)} on series  lengths: 3, 5, 10 in three different integer ranges evaluated as Interpolation (IN) and Extrapolation (EX) exact-match scores on 1K test data.}
\label{tab:sort}
\end{table*}

\section{Results and Error Analysis}

\begin{figure}
  \includegraphics[width=\columnwidth]{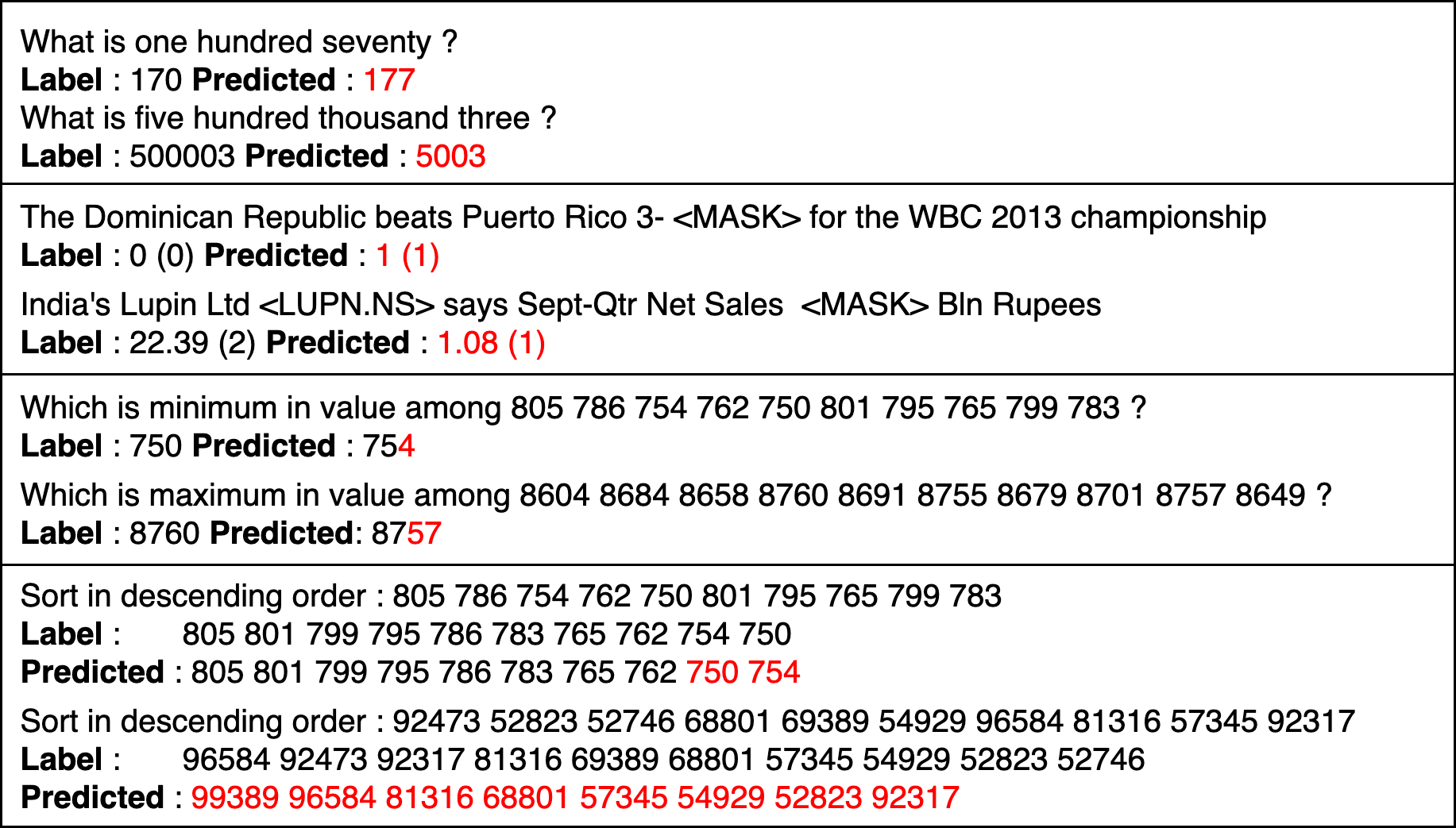}
  \caption{Two incorrect predictions for each  task.}
  \label{fig:T5_errors}
\end{figure}

Table \ref{tab:decoding} shows, all versions of T5 benefit when they are trained with split representation. When trained with 4.9K data, T5-SM gains 24\% points in interpolation evaluation where T5-LG gains only 2\%. None of the models perform well on unseen number data  ranges. In fewer shot interpolation settings however, only the T5-LG model maintains its performance beyond 90\% which is not surprising because of its large parameter-space. We noticed that the best model could only partially decode numbers having multiple zeros (Figure \ref{fig:T5_errors}). In the first example, the model predicts an extra seven and in the second (extrapolation), it ignored the key word `hundred' as it attempts to fit this unseen data into a similar seen number range (4 digits).

In magnitude order prediction (Table \ref{tab:magnitude}), T5-LG's performance   improves  by 5 $\mu$F1  in article title. 
For extrapolation  (Table \ref{tab:magnitude_cross}), all T5 versions beats previous estimates (BiGRU) by at most 25\%. This shows that  T5 can learn robust numeric representations based on contexts. 
Both the samples in Fig \ref{fig:T5_errors} are hard as they need prior explicit knowledge. Yet they are able to predict numbers in similar feasible ranges. 
This shows that the model is not randomly assigning magnitude but has learnt based on the domain and context. 
We found  that, the best T5 model predicted an order of 1 instead of 2 for market and article data making a maximum error of 39.07\% and 33.59\% respectively.

Table \ref{tab:minmax} shows List-MinMax results. Both T5-BS and T5-LG  perform  over 80\% across all ranges and series lengths. T5-SM however, degrades in performance as the range increases along with the list size. As the model learns more variations in numbers, the extrapolation performance increases to a max of 81\% (List-Min) and 84.9\% (List-Max). But the performance drops as series length increases. The best model predicted second minimum and maximum element in the examples of Fig \ref{fig:T5_errors}.

From the sorting results (Table \ref{tab:sort}), we see T5-SM performance drops (18-22\% from 2-3 digits, 8-9\% from 3-4) as number ranges increase across series length of 3. 
T5-SM fails to generate a single correct order for a series of 10 elements and achieves less than 10\% success in 5-element series across all ranges. This degrading performance can be attributed to its mere 60M parameter space.
As the number of parameters keep increasing 
the models performs consistently across each of 3, 5, 10 elements in series, both for interpolation and extrapolation settings.
With the increasing  range of training data, the models become more robust to extrapolated numbers across all series lengths with 8-30\% change in ascending order and 7-20\% change in descending order. Finally, for sorting, we find a variety of incorrect predictions: missing order of one element, omission of one and two elements or repeating a particular element. 

\begin{figure}
  \includegraphics[width=\columnwidth]{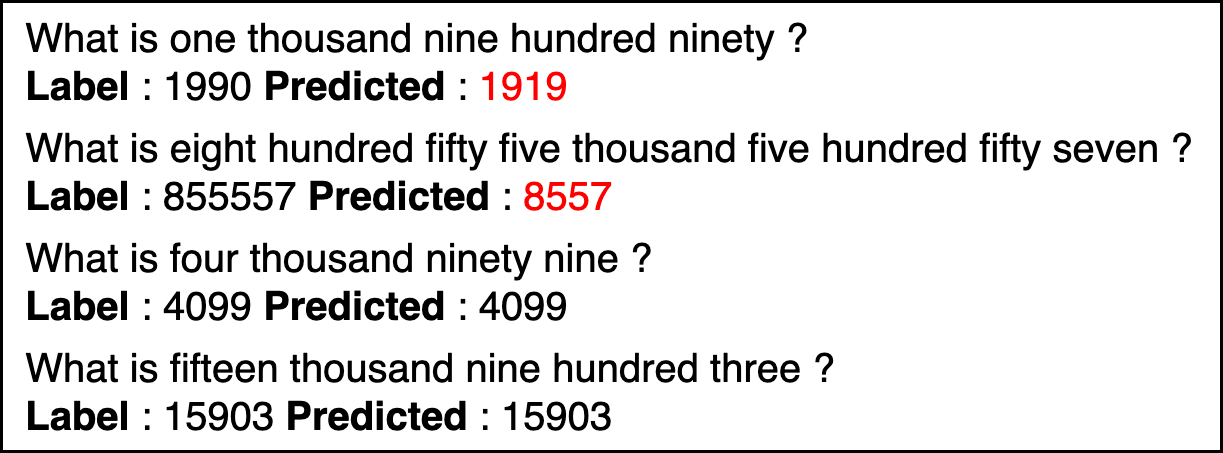}
  \caption{Some predictions for Numeration task.}
  \label{fig:2_T5_errors_num}
\end{figure}

\begin{figure}
  \includegraphics[width=\columnwidth]{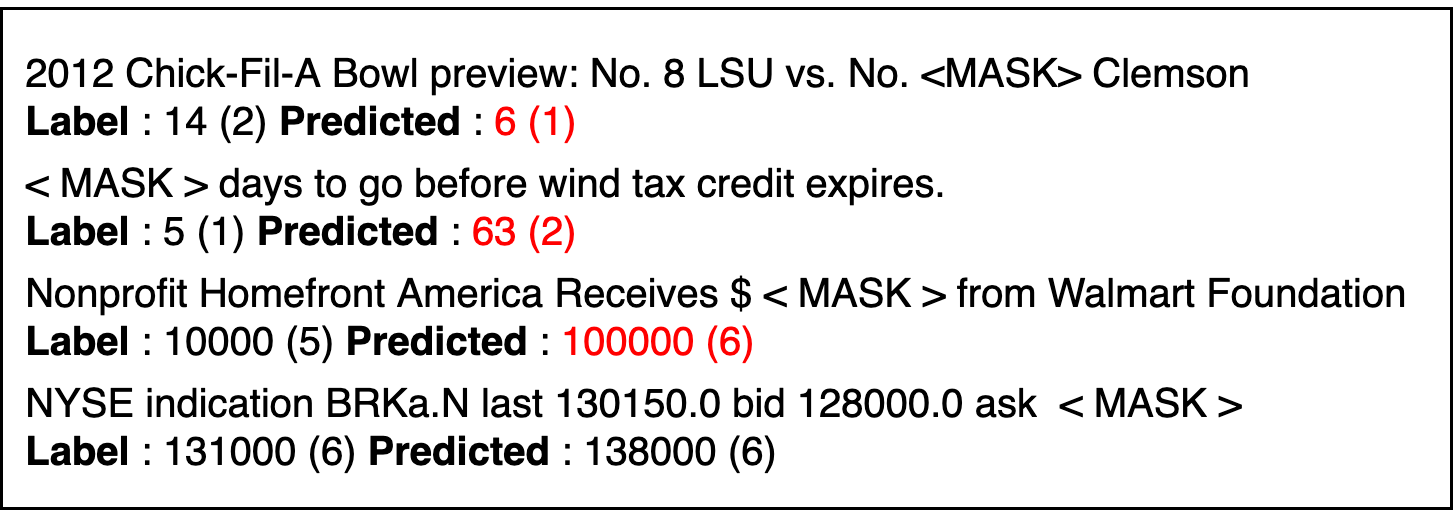}
  \caption{Magnitude Order Prediction Examples.}
  \label{fig:2_T5_errors_MOP}
\end{figure}

\begin{figure}
  \includegraphics[width=\columnwidth]{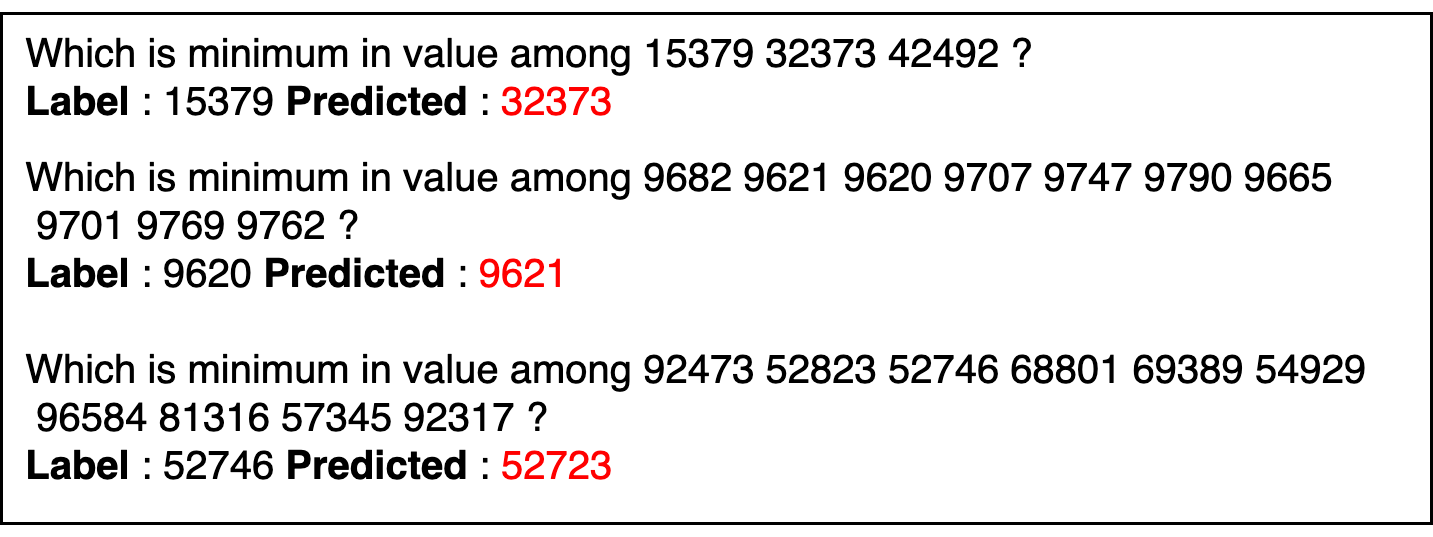}
  \caption{Some predictions for List-MinMax task.}
  \label{fig:2_T5_errors_minmax}
\end{figure}

\begin{figure}
  \includegraphics[width=\columnwidth]{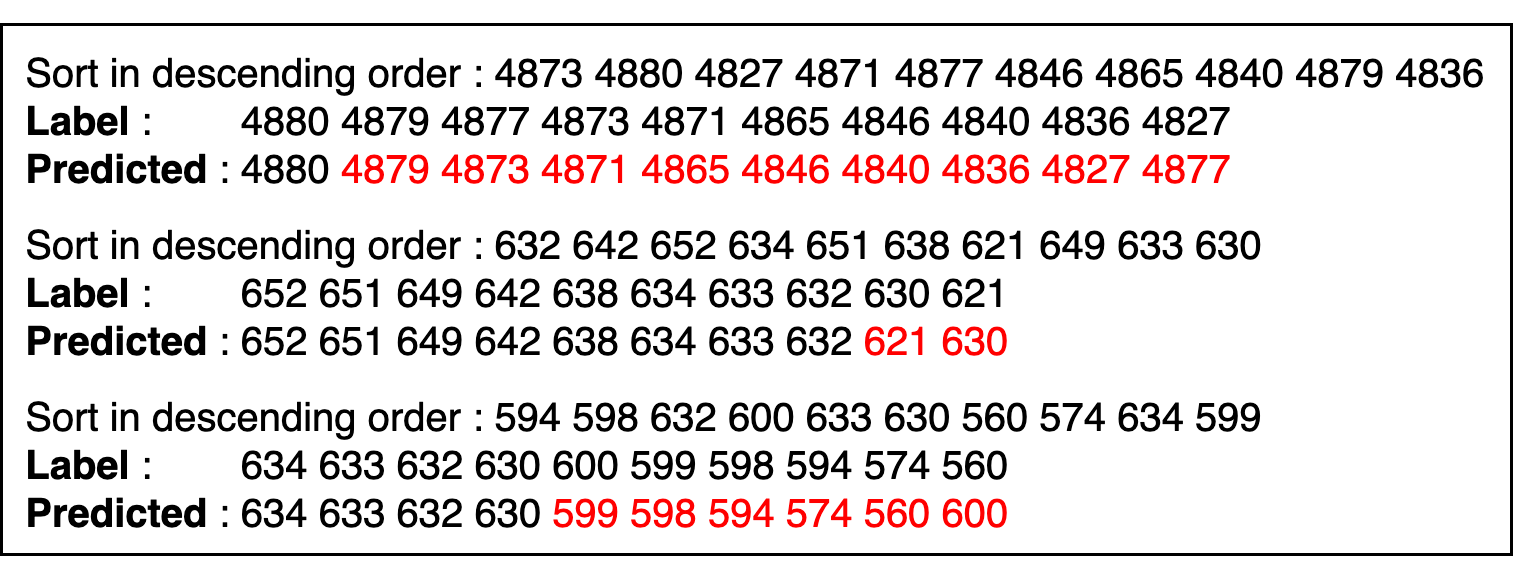}
  \caption{Some predictions for List-Sort task.}
  \label{fig:2_T5_errors_sort}
\end{figure}

Overall, none of the models were able to perform well on extrapolation samples showing the inherent rules of numeracy is  difficult for these models to learn. But, it also shows,  more variations in numbers (increasing the range) help them perform better in extrapolation setting. The smaller model's limited parameter-space affects its performance in all four tasks whereas larger models are able to pick up some numeracy skills through training. We show more predictions in Figure \ref{fig:2_T5_errors_num}, \ref{fig:2_T5_errors_MOP}, \ref{fig:2_T5_errors_minmax}, \ref{fig:2_T5_errors_sort}.


\textbf{Analysis of NT5:}
We test with the NT5 \cite{yang2021nt5} model on all our experiments and compared the results with T5-small. For the Numeration task with the split number representation NT5 performed 73.07 (accuracy), a 4\% improvement over T5. The performance however did not improve for the MinMax and Sorting tasks. For 3-element sorting it dropped by 10-20\%.  In the Magnitude Order Prediction, we find the cross-domain (extrapolation) $\mu$F1 score increases by 5-7\% while in-domain decreases by 3-6\%. This might be because NT5 has seen more variety of contexts of numbers and can generalize well on this task.

\section{Related Works}

\textbf{Numeracy Tests:} Multiple numeracy tests have been proposed to evaluate the static word embeddings \cite{naik2019exploring}  like GloVe, Word2Vec, FastText and contextual embeddings \cite{wallace-etal-2019-nlp} like BERT through probing tasks like numeration, magnitude comparison, addition, list-maximum. Multilingual numeration \cite{johnson-multilingual} tests have been performed by probing models like DistilBERT, XLM, and BERT. CNN, BiGRU models have been shown to perform well in magnitude order prediction \cite{chen2019numeracy} and T5 on addition and subtraction tasks \cite{nogueira2021investigating} through training on similar texts. We, however focus on studying how much text-to-text transfer models (T5) can learn across four fundamental numeracy tasks in samples containing both in-domain and out-of-domain numerical ranges.

\noindent
\textbf{Specially Designed Models:}  NALU \cite{trasknalu}, NAU and NMU \cite{madsen2020neural}, numBERT \cite{zhang-etal-2020-language-embeddings}, GenBERT \cite{genbert}, NT5 \cite{yang2021nt5} have emerged in the last few years to incorporate arithmetic skills into models through specially designed architecture or fine-tuning tasks which improves the performance in synthetic arithmetic or  crowd-sourced numerical reasoning tasks like DROP.

\noindent
\textbf{Numerical Embeddings:} 
There are limited prior works in numeracy aware embeddings which show good performance in extrapolation setting.
One approach \cite{jiang2019learning} represents numerals as a weighted average of  prototype numeral embeddings obtained using either self organizing map or Gaussian Mixture models. DICE \cite{sundararaman2020methods} is a deterministic numeral embedding approach, independent of corpus, which preserves the relative magnitude between two numerals and their embeddings.













\section{Conclusion \& Future Works}
We show that  text-to-text models  are able to learn numeracy quite well in an interpolation setting. Our extensive experiments show that T5 models struggle to learn with numbers outside training data ranges. 
We believe that, to make further progress in transfer learning, models need to achieve such elementary numeracy skills and this gap between interpolation and extrapolation performance needs to be reduced. We are of the opinion that, adding more data would not bridge this gap since domain of numbers is open. However, special pre-training objectives for digits rather than whole numbers can be designed to teach the inherent numeracy to models. In future, we intend to explore these objectives centered around preserving numeracy rules in transfer-learned models to generalize between in-domain and out-of-domain numbers.

\section*{Acknowledgement}
The authors acknowledge support from DARPA grant number FA875019C0003 for this project.

\section*{Ethical Considerations}
In this paper, we analyze  performance of three publicly available T5 models on four numeracy tasks. For Magnitude Order Prediction task we use publicly available dataset, Numeracy600K. We synthetically create the data for rest of the tasks. 

\bibliography{anthology,custom}

\begin{thebibliography}{19}
\expandafter\ifx\csname natexlab\endcsname\relax\def\natexlab#1{#1}\fi

\bibitem[{Chen et~al.(2019)Chen, Huang, Takamura, and Chen}]{chen2019numeracy}
Chung-Chi Chen, Hen-Hsen Huang, Hiroya Takamura, and Hsin-Hsi Chen. 2019.
\newblock Numeracy-600k: learning numeracy for detecting exaggerated
  information in market comments.
\newblock In \emph{Proceedings of the 57th Annual Meeting of the Association
  for Computational Linguistics}, pages 6307--6313.

\bibitem[{Dua et~al.(2019)Dua, Wang, Dasigi, Stanovsky, Singh, and
  Gardner}]{dua-etal-2019-drop}
Dheeru Dua, Yizhong Wang, Pradeep Dasigi, Gabriel Stanovsky, Sameer Singh, and
  Matt Gardner. 2019.
\newblock Drop: A reading comprehension benchmark requiring discrete reasoning
  over paragraphs.
\newblock \emph{arXiv preprint arXiv:1903.00161}.

\bibitem[{Geva et~al.(2020)Geva, Gupta, and Berant}]{genbert}
Mor Geva, Ankit Gupta, and Jonathan Berant. 2020.
\newblock \href {https://doi.org/10.18653/v1/2020.acl-main.89} {Injecting
  numerical reasoning skills into language models}.
\newblock In \emph{Proceedings of the 58th Annual Meeting of the Association
  for Computational Linguistics}, pages 946--958, Online. Association for
  Computational Linguistics.

\bibitem[{Jiang et~al.(2019)Jiang, Nian, Guo, Chu, Zhao, Shen, and
  Tu}]{jiang2019learning}
Chengyue Jiang, Zhonglin Nian, Kaihao Guo, Shanbo Chu, Yinggong Zhao, Libin
  Shen, and Kewei Tu. 2019.
\newblock Learning numeral embeddings.
\newblock \emph{arXiv preprint arXiv:2001.00003}.

\bibitem[{Johnson et~al.(2020)Johnson, Mak, Barker, and
  Loessberg-Zahl}]{johnson-multilingual}
Devin Johnson, Denise Mak, Andrew Barker, and Lexi Loessberg-Zahl. 2020.
\newblock \href {https://doi.org/10.18653/v1/2020.blackboxnlp-1.18} {Probing
  for multilingual numerical understanding in transformer-based language
  models}.
\newblock In \emph{Proceedings of the Third BlackboxNLP Workshop on Analyzing
  and Interpreting Neural Networks for NLP}, pages 184--192, Online.
  Association for Computational Linguistics.

\bibitem[{Lin et~al.(2020)Lin, Lee, Khanna, and Ren}]{lin2020numersense}
Bill~Yuchen Lin, Seyeon Lee, Rahul Khanna, and Xiang Ren. 2020.
\newblock Birds have four legs?! numersense: Probing numerical commonsense
  knowledge of pre-trained language models.
\newblock In \emph{Proceedings of EMNLP}.
\newblock To appear.

\bibitem[{Liu et~al.(2021)Liu, Yuan, Fu, Jiang, Hayashi, and
  Neubig}]{liu2021pretrain}
Pengfei Liu, Weizhe Yuan, Jinlan Fu, Zhengbao Jiang, Hiroaki Hayashi, and
  Graham Neubig. 2021.
\newblock \href {http://arxiv.org/abs/2107.13586} {Pre-train, prompt, and
  predict: A systematic survey of prompting methods in natural language
  processing}.

\bibitem[{Madsen and Johansen(2020)}]{madsen2020neural}
Andreas Madsen and Alexander~Rosenberg Johansen. 2020.
\newblock Neural arithmetic units.
\newblock \emph{arXiv preprint arXiv:2001.05016}.

\bibitem[{Mishra et~al.(2020)Mishra, Mitra, Varshney, Sachdeva, and
  Baral}]{mishra2020towards}
Swaroop Mishra, Arindam Mitra, Neeraj Varshney, Bhavdeep Sachdeva, and Chitta
  Baral. 2020.
\newblock Towards question format independent numerical reasoning: A set of
  prerequisite tasks.
\newblock \emph{arXiv preprint arXiv:2005.08516}.

\bibitem[{Naik et~al.(2019)Naik, Ravichander, Rose, and
  Hovy}]{naik2019exploring}
Aakanksha Naik, Abhilasha Ravichander, Carolyn Rose, and Eduard Hovy. 2019.
\newblock Exploring numeracy in word embeddings.
\newblock In \emph{Proceedings of the 57th Annual Meeting of the Association
  for Computational Linguistics}, pages 3374--3380.

\bibitem[{Nogueira et~al.(2021)Nogueira, Jiang, and
  Li}]{nogueira2021investigating}
Rodrigo Nogueira, Zhiying Jiang, and Jimmy Li. 2021.
\newblock Investigating the limitations of the transformers with simple
  arithmetic tasks.
\newblock \emph{arXiv preprint arXiv:2102.13019}.

\bibitem[{Raffel et~al.(2020)Raffel, Shazeer, Roberts, Lee, Narang, Matena,
  Zhou, Li, and Liu}]{T5}
Colin Raffel, Noam Shazeer, Adam Roberts, Katherine Lee, Sharan Narang, Michael
  Matena, Yanqi Zhou, Wei Li, and Peter~J. Liu. 2020.
\newblock \href {http://jmlr.org/papers/v21/20-074.html} {Exploring the limits
  of transfer learning with a unified text-to-text transformer}.
\newblock \emph{Journal of Machine Learning Research}, 21(140):1--67.

\bibitem[{Rajpurkar et~al.(2016)Rajpurkar, Zhang, Lopyrev, and
  Liang}]{rajpurkar-etal-2016-squad}
Pranav Rajpurkar, Jian Zhang, Konstantin Lopyrev, and Percy Liang. 2016.
\newblock \href {https://doi.org/10.18653/v1/D16-1264} {{SQ}u{AD}: 100,000+
  questions for machine comprehension of text}.
\newblock In \emph{Proceedings of the 2016 Conference on Empirical Methods in
  Natural Language Processing}, pages 2383--2392, Austin, Texas. Association
  for Computational Linguistics.

\bibitem[{Sundararaman et~al.(2020)Sundararaman, Si, Subramanian, Wang,
  Hazarika, and Carin}]{sundararaman2020methods}
Dhanasekar Sundararaman, Shijing Si, Vivek Subramanian, Guoyin Wang, Devamanyu
  Hazarika, and Lawrence Carin. 2020.
\newblock Methods for numeracy-preserving word embeddings.
\newblock In \emph{Proceedings of the 2020 Conference on Empirical Methods in
  Natural Language Processing (EMNLP)}, pages 4742--4753.

\bibitem[{Thawani et~al.(2021)Thawani, Pujara, Szekely, and
  Ilievski}]{thawani2021representing}
Avijit Thawani, Jay Pujara, Pedro~A Szekely, and Filip Ilievski. 2021.
\newblock Representing numbers in nlp: a survey and a vision.
\newblock \emph{arXiv preprint arXiv:2103.13136}.

\bibitem[{Trask et~al.(2018)Trask, Hill, Reed, Rae, Dyer, and
  Blunsom}]{trasknalu}
Andrew Trask, Felix Hill, Scott~E Reed, Jack Rae, Chris Dyer, and Phil Blunsom.
  2018.
\newblock \href
  {https://proceedings.neurips.cc/paper/2018/file/0e64a7b00c83e3d22ce6b3acf2c582b6-Paper.pdf}
  {Neural arithmetic logic units}.
\newblock In \emph{Advances in Neural Information Processing Systems},
  volume~31. Curran Associates, Inc.

\bibitem[{Wallace et~al.(2019)Wallace, Wang, Li, Singh, and
  Gardner}]{wallace-etal-2019-nlp}
Eric Wallace, Yizhong Wang, Sujian Li, Sameer Singh, and Matt Gardner. 2019.
\newblock \href {https://doi.org/10.18653/v1/D19-1534} {Do {NLP} models know
  numbers? probing numeracy in embeddings}.
\newblock In \emph{Proceedings of the 2019 Conference on Empirical Methods in
  Natural Language Processing and the 9th International Joint Conference on
  Natural Language Processing (EMNLP-IJCNLP)}, pages 5307--5315, Hong Kong,
  China. Association for Computational Linguistics.

\bibitem[{Yang et~al.(2021)Yang, Chen, Chen, and Cer}]{yang2021nt5}
Peng-Jian Yang, Ying~Ting Chen, Yuechan Chen, and Daniel Cer. 2021.
\newblock Nt5?! training t5 to perform numerical reasoning.
\newblock \emph{arXiv preprint arXiv:2104.07307}.

\bibitem[{Zhang et~al.(2020)Zhang, Ramachandran, Tenney, Elazar, and
  Roth}]{zhang-etal-2020-language-embeddings}
Xikun Zhang, Deepak Ramachandran, Ian Tenney, Yanai Elazar, and Dan Roth. 2020.
\newblock \href {https://doi.org/10.18653/v1/2020.findings-emnlp.439} {Do
  language embeddings capture scales?}
\newblock In \emph{Findings of the Association for Computational Linguistics:
  EMNLP 2020}, pages 4889--4896, Online. Association for Computational
  Linguistics.

\end{thebibliography}
\bibliographystyle{acl_natbib}

\appendix

\section{Appendix}

\subsection{Data Statistics Experimental Setup}
\textbf{Numeration:} We have  4906, 2097, 2997 in train, dev and test  respectively. We make sure that all  numbers within 10K are present in any of train, dev or test. For extrapolation we select 1K integers randomly from every 10K range from  [10K,1000K) making it a total of 99K.

\noindent
\textbf{Magnitude Order Prediction:}
For this data we consider 450K, 50K and 100K samples for train, dev and test data respectively from each of market comments and article titles data.

\noindent
\textbf{List-Sort:}
We consider both the task of arranging in ascending and descending orders since if a series is already sorted in ascending order the model can directly predict by copying it from the given input.

\begin{figure}[!htb]
  \includegraphics[width=\columnwidth]{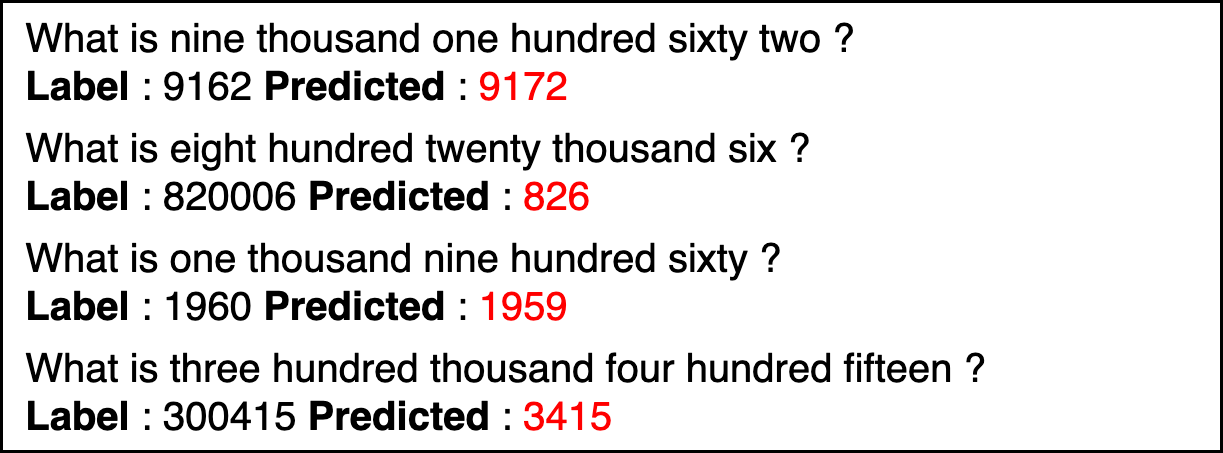}
  \caption{More predictions for Numeration task.}
  \label{fig:T5_errors_num}
\end{figure}

\subsection{Hyperparameters}
For all the experiments we use maximum sequence length of 128 and 256 for question context. The  maximum sequence length of the answers is kept as [5, 10, 20, 25] for different tasks. We ran for 20 epochs and  save a model based on validation EM performance. Our training and validation batch size varies between [2, 4, 8, 16, 32] based on the experiment. We work on 4 Tesla V100 GPUs. We use AdamW optimizer and StepLR scheduler with step size of 2, learning rate of 5e-5 and gamma of 0.1.

\begin{figure}[!htb]
  \includegraphics[width=\columnwidth]{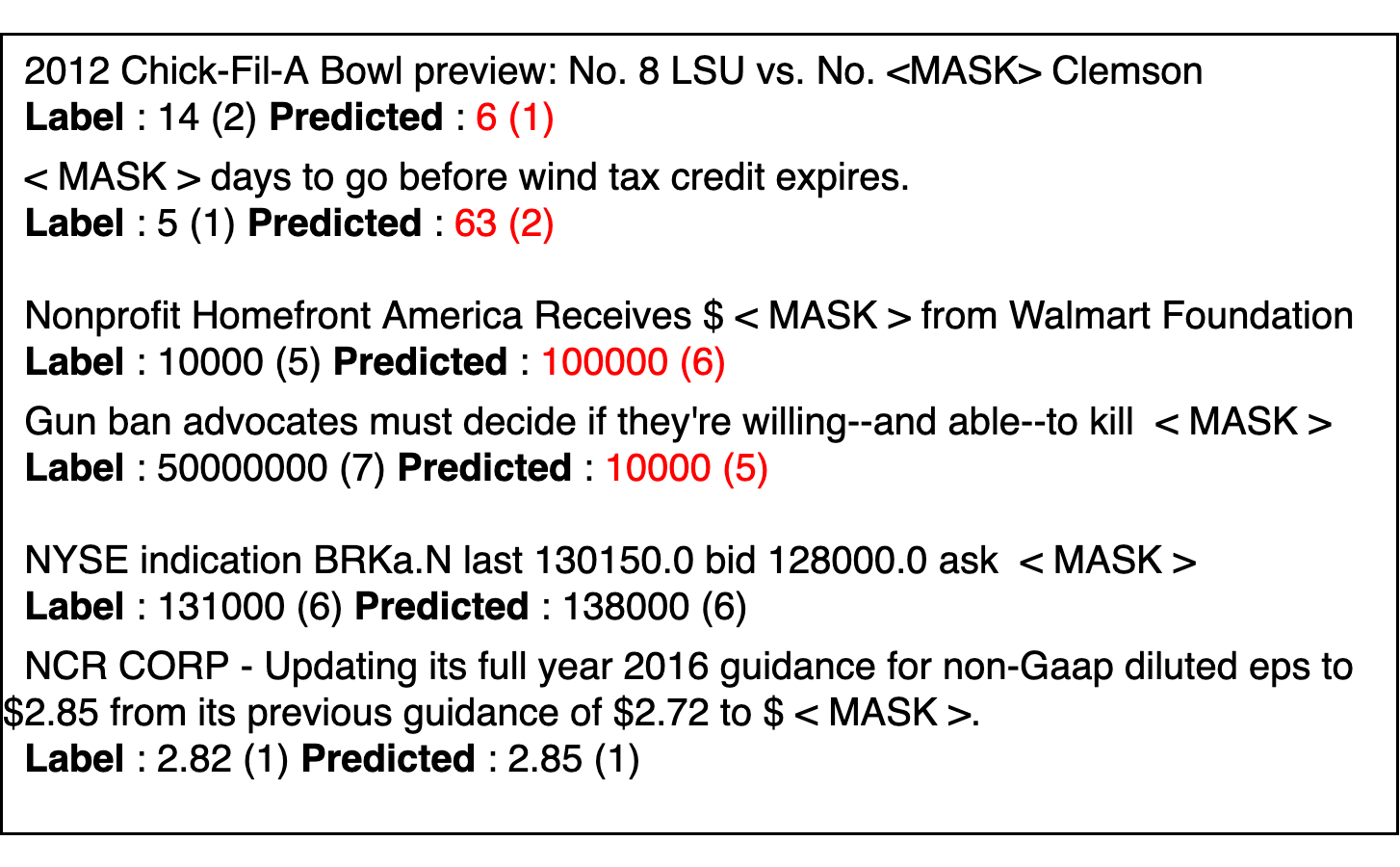}
  \caption{More Magnitude Order Prediction Examples.}
  \label{fig:T5_errors_MOP}
\end{figure}

\subsection{Results and Error analysis}

\begin{figure}
  \includegraphics[width=\columnwidth]{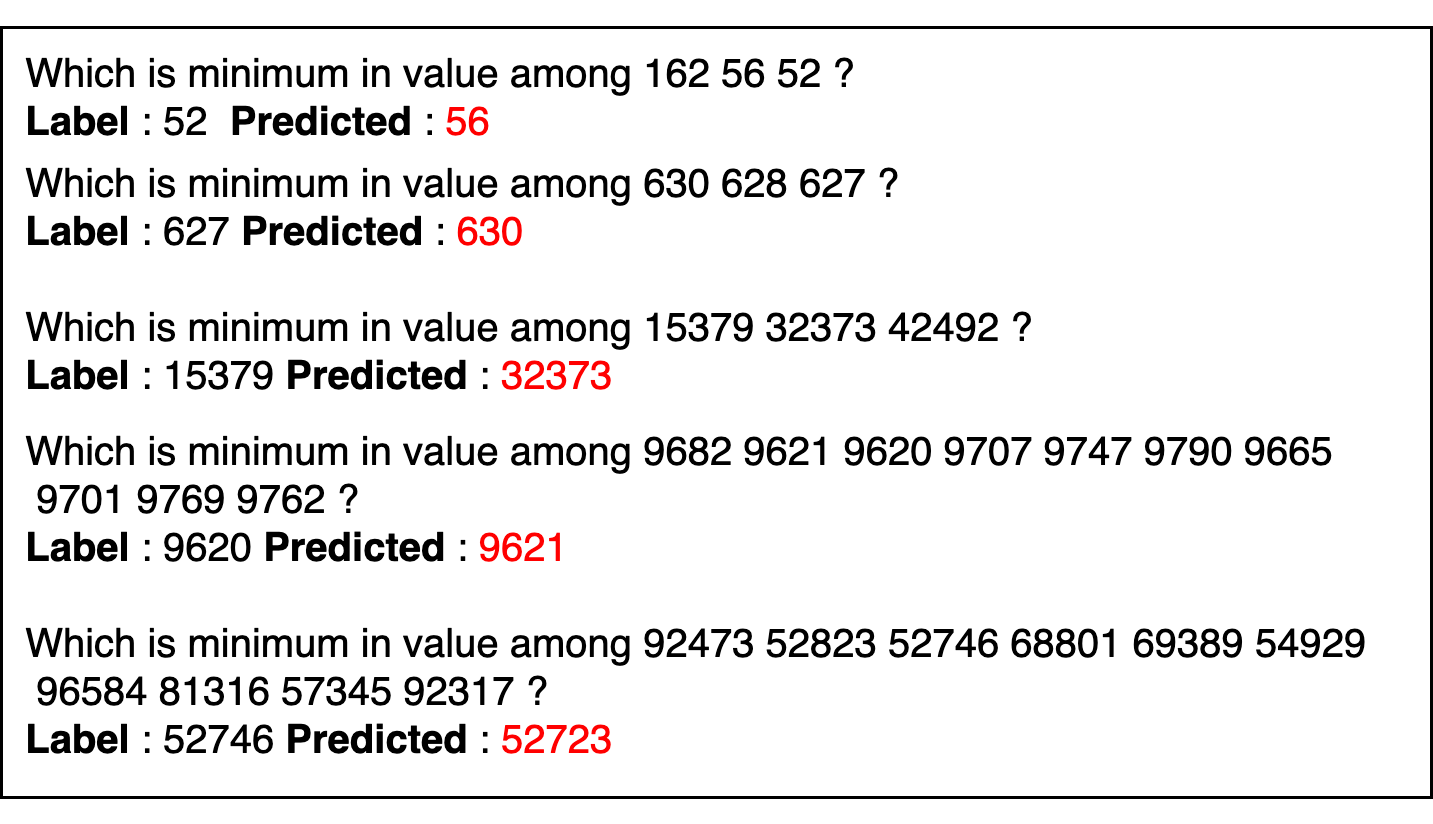}
  \caption{More predictions for List-MinMax task.}
  \label{fig:T5_errors_minmax}
\end{figure}

\begin{figure}
  \includegraphics[width=\columnwidth]{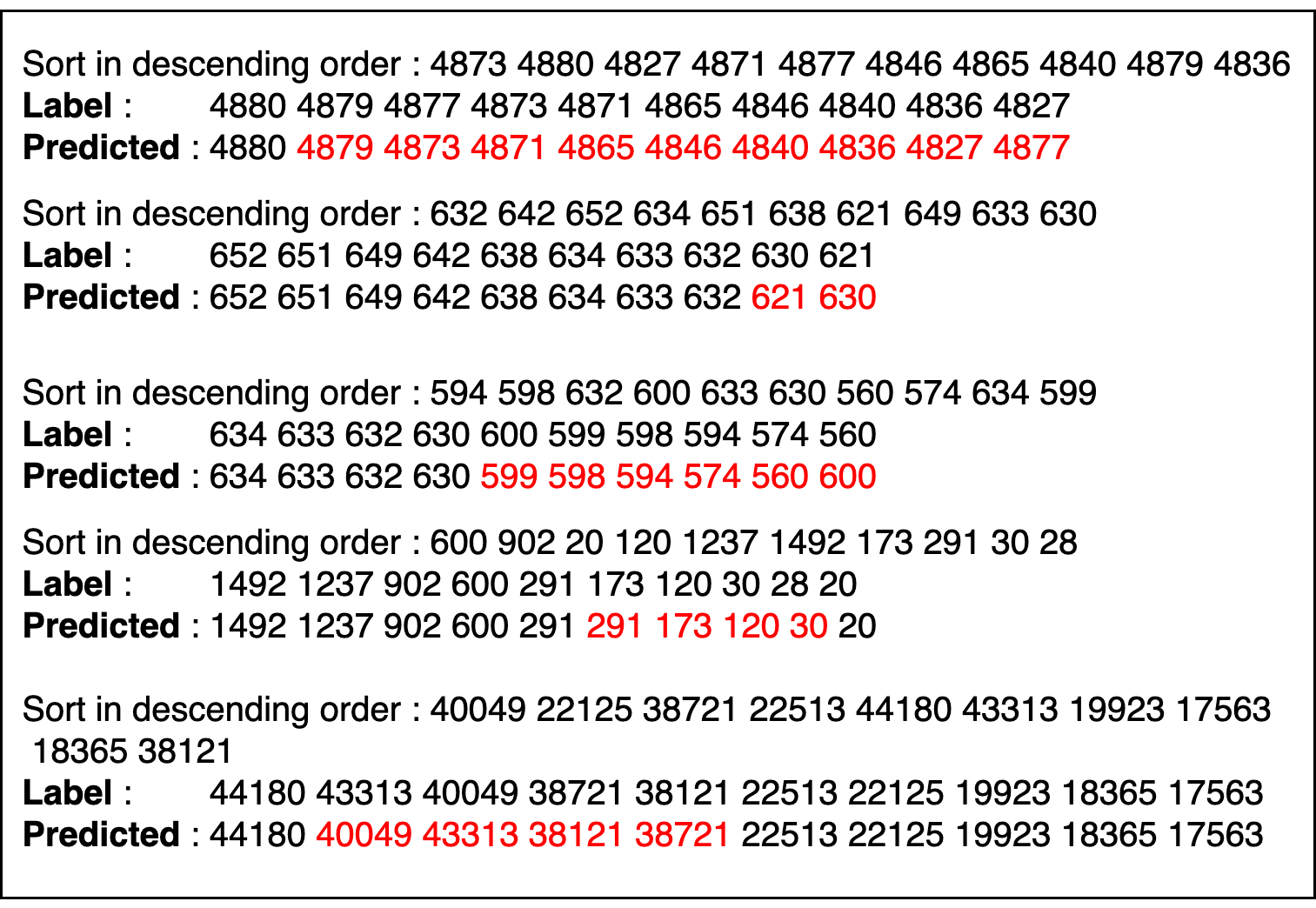}
  \caption{More predictions for List-Sort task.}
  \label{fig:T5_errors_sort}
\end{figure}

\textbf{Magnitude Order Prediction:}
We also experimented with zero-shot magnitude order predictions. We found 553 and 8783 exact-matches out of 100K test data using T5-large which shows that the performance is very poor without proper fine-tuning. 
We show some more predictions of the best performing T5 model in Figure \ref{fig:T5_errors_num}, \ref{fig:T5_errors_MOP}, \ref{fig:T5_errors_minmax}, \ref{fig:T5_errors_sort}.

\end{document}